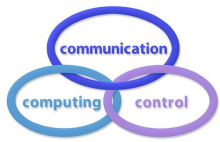
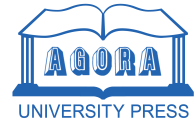

# Weighted Random Search for CNN Hyperparameter Optimization

R. Andonie, A.C. Florea

**Razvan Andonie**
1. Department of Computer Science
Central Washington University, USA
2. Department of Electronics and Computers
Transilvania University of Braşov, Romania
*Corresponding author: andonie@cwu.edu

**Adrian-Cătălin Florea**
Department of Electronics and Computers
Transilvania University of Braşov, Romania
acflorea@unitbv.ro

**Abstract**

Nearly all model algorithms used in machine learning use two different sets of parameters: the training parameters and the meta-parameters (hyperparameters). While the training parameters are learned during the training phase, the values of the hyperparameters have to be specified before learning starts. For a given dataset, we would like to find the optimal combination of hyperparameter values, in a reasonable amount of time. This is a challenging task because of its computational complexity. In previous work [11], we introduced the Weighted Random Search (WRS) method, a combination of Random Search (RS) and probabilistic greedy heuristic. In the current paper, we compare the WRS method with several state-of-the art hyperparameter optimization methods with respect to Convolutional Neural Network (CNN) hyperparameter optimization. The criterion is the classification accuracy achieved within the same number of tested combinations of hyperparameter values. According to our experiments, the WRS algorithm outperforms the other methods.

## 1 Introduction

Machine learning models usually have two types of parameters: the training parameters and hyperparameters. The training parameters are learned during the training phase, whereas the values of the hyperparameters have to be specified before learning starts. For instance, the hyperparameters of neural networks typically specify the architecture of the network (number and type of layers, number and type of nodes, etc). For a given dataset, we would like to find the optimal combination of hyperparameter values, in a reasonable amount of time. This process is called *hyperparameter optimization*.

Formally, the aim of hyperparameter optimization is to find the hyperparameters of a given model that return the best performance as measured on a validation set. This optimization can be represented as [3]:

$$x^* = arg \min_{x \in \aleph} f(x), \tag{1}$$

where $f(x)$ is the objective function to minimize (such as RMSE or error rate) evaluated on the validation set; $x^*$ is the set of hyperparameters that yields the lowest value of the score, and $x$ can take on any value in the domain $\aleph$. In simple terms, we want to find the model hyperparameters that yield the best score on the validation set metric.

More detailed, eq (1), can be written as:

$$x^* = arg \min_{x \in \aleph} f(x, \gamma^*; S_{validation}) \tag{2}$$

Eq (2) includes an inner optimization used to find $\gamma^*$, the optimal value of $\gamma$, for the current $x$ value:

$$\gamma^* = arg \min_{\gamma \in \Gamma} f(x, \gamma; S_{train}) \tag{3}$$

where $S_{validation}$ and $S_{train}$ denote the validation and training datasets respectively; $\gamma$ is the set of learned parameters in the domain $\Gamma$ through minimization of the training error.

Eq (1) refers to a "validation set", without further details. We have to evaluate the expectation of the score over an unknown distribution and we usually approximate this expectation using a three-way split, dividing the dataset into a training, validation, and test dataset. Having a training-validation pair for hyperparameter tuning allows us to keep the test set "independent" for model evaluation. We will use three-way split in our implementations and experiments, but we will omit such details in our presentation, since they are standard.

The validation process can be also generalized at the *model* level. In this case, at meta-level we have several possible models (i.e., feedforward neural network, random forest, support vector machine, etc), each with its own set of hyperparameters. The process of finding the best-performing model from a set of models that were produced by different hyperparameter settings is called *model selection* [32]. In this case, we search for the best model and, within the model, for the best combination of hyperparameter values. Many optimizers attempt to optimize both the choice of the model and the hyperparameters of the model: Auto-WEKA, Hyperopt-sklearn, AutoML, auto-sklearn, etc. For simplicity, we will consider here only optimization of hyperparameters within the framework of the same model.

Evaluating the objective function to find the score can be computationally very expensive. Each time we try different hyperparameters, we typically re-train the model, make predictions on the validation data, and then calculate the validation metric. We call this re-training + evaluation for one combination of hyperparameter values a *trial*.

Lately, there has been high interest in the area of hyperparameter optimization, related to the importance and complexity of the deep learning architectures. The challenge is that in deep learning, we typically deal with large training sets. In addition, we may also encounter high-dimensional search spaces, which according to the curse of dimensionality principle have to be covered by an exponential increasing number of points (in this case, combinations of hyperparameters). Therefore, deep learning hyperparameter optimization algorithms are computationally intensive. For instance, to optimize a deep learning architecture for the CIFAR-10 and ImageNet datasets required 2,000 GPU days of reinforcement learning [43] or 3,150 GPU days of evolution [33].

The performance of a hyperparameter optimization method is determined by several factors:

- F1. The execution time of each trial. This time depends on the size and the dimensionality of the training set.

- F2. The number of evaluated combinations of hyperparameters (the number of trials).

- F3. The performance of the search procedure. Within the same number of trials, different optimization methods can achieve different scores, depending on how "smart" they are. This depends on which combinations of hyperparameter values are tested and in which order.

We restrict the discussion to F3 and our the focus is to increase the score of a classifier within the same computational budget. In [11] we introduced the WRS method, an improved RS method which can be used in hyperparameter optimization. Instead of a blind RS search, the WRS method uses information from previous trials to guide the search process toward next interesting trials. WRS combines RS and a probabilistic greedy heuristic. We formally proved that the WRS method finds the global optimum faster than RS: on average, WRS converges faster than RS. We also proved that WRS is convergent. Experimentally, for the same number of trials, the WRS algorithm produced significantly better results than RS on the CIFAR-10 dataset[1]. So far, the WRS method was never compared with other optimization methods, except RS.

---

[1] http://www.cs.toronto.edu/ kriz/cifar.html

In the current contribution, we compare the WRS method with several state-of-the-art hyperparameter optimization methods. Since CNNs are presently by far the most popular deep neural architecture, our experiments focus on this architecture. We compare the classification accuracy achieved within the same number of trials. As a result, the WRS optimization algorithm proves to be competitive, even when compared with some of the best known methods.

The paper proceeds as follows. Section 2 presents the state-of-the-art of CNN hyperparameter optimization methods, whereas Section 3 is a brief overview of the current software implementing these methods. Section 4 describes in an intuitive way the WRS algorithm. Our main contribution is Section 5, where we present experimental results, comparing WRS with several state-of-the-art methods. The paper is then concluded in Section 6.

## 2 Related work: Methods for CNN Hyperparameter Optimization

To set the stage for WRS, we first introduce the currently most used methods for hyperparameter optimization, with focus on CNN architectures.

An overview of fundamental methods for hyperparameter optimization can be found, for instance, in [4, 21]. The best known methods are: Grid Search (GS), Random Search (RS), Bayesian Optimization (BO), Nelder-Mead (NM), Simulated Annealing (SA), Particle Swarm Optimization, and Evolutionary Algorithms. Until recently, the most commonly used hyperparameter optimization strategy was a combination of GS and manual tuning. More complex techniques are now largely available in a variety of software packages.

RS consists in drawing samples from the parameter space following a particular distribution for each of the parameters. Using the same number of trials, RS generally yields better results than GS or more complicated hyperparameter optimization methods. Especially in higher dimensional spaces, the computation resources required by RS methods are significantly lower than for GS [22]. Recent attempts to optimize the RS algorithm are: Li's *et al.* Hyperband [23], Domhan *et al.* [9], and Florea *et al.* [10], where we introduced a dynamically computed stopping criterion for RS, reducing the number of trials without reducing the generalization performance.

By evaluating hyperparameters that appear more promising from past results, Bayesian methods can find better model settings than RS or GS in fewer iterations. A reviews of BO can be found in [35]. Sequential Model-Based Optimization (SMBO) methods are a formalization of BO. The sequential optimization refers to running trials one after another, each time trying better hyperparameters and updating a probability model (a surrogate). There are several variants of SMBO methods, based on Gaussian Processes, Random Forest Regression, Tree-structured Parzen Estimators (TPE), etc.

In low-dimensional problems with numerical hyperparameters, the best available hyperparameter optimization methods use BO [36]. However, BO is restricted to problems of moderate dimension [35].

In current CNN architectures, the number of hyperparameters has increased significantly. These hyperparameters include: number of layers, filter size, number of feature maps, stride, pooling regions and pooling sizes, and the number of units in each fully-connected layer. The number of these parameters can be the in the order of tenths or even hundreds. Not only must a hyperparameter optimization algorithm optimize over variables which are discrete, ordinal, and continuous, but it must simultaneously choose which variables to optimize - a difficult task. Currently, no work covers optimization of every hyperparameter in designing a CNN architecture [1]. We briefly refer in the following to some recent CNN hyperparameter optimization results.

The SMBO was recently used for CNN hyperparameter optimization [39]. By learning the hyperparameter response function, SMBO can pick the next promising hyperparameter combination. In [36], BO with Gaussian Processes are used to optimize nine hyperparameters of a CNN, including the learning rate, epochs, the four weights of the CNN layers, and the parameters of response contrast normalization (width, scale, and power). Another approach of this type is [16].

Evolutionary approaches for CNN optimization approaches can be found in [28, 41].

In high-dimensional hyperparameter optimization, tree-based models are presently considered more successful: TPE [4] and Random Forests [14]. Very recent deep learning hyperparameter optimization techniques are [17], [25], and [9]. These techniques attempt to reduce as much as possible the number of evaluated combinations of hyperparameter values. However, for large training sets, even if the number of evaluated combinations is reduced, the total training time increases very much, becoming the complexity bottleneck.

Hinz *et al.* applied several algorithms to optimize the architecture of CNNs for image inputs with increasing resolution [13]. That way, they conceptually used the same input data, but in different input dimensions. They found that the same hyperparameters are important for a given dataset, independent of the image resolution. Furthermore, the optimal value for most hyperparameters also seems to be independent of the image resolution. Consequently, they proposed an approach to speed up the hyperparameter optimization procedure by starting on smaller images and increasing the resolution during the optimization procedure.

Albelwi *at al.* recently proposed a general method for CNN hyperparameter optimization [2]. Their approach is based on *a)* a preprocessing step, when they reduce the training set using instance selection; *b)* the NM optimization method. For the CIFAR-10, CIFAR-100, and MINIST datasets, the following seven hyperparameters (each with its corresponding range of possible values) were considered: depth, number of fully-connected layers, number of filters, kernel sizes, number of pooling layers, pooling region sizes, number of neurons in fully-connected layers.

Other recent deep learning hyperparameter optimization techniques are:

- Deterministic RBF surrogates [16].

- Covariance Matrix Adaptation Evolution Strategy [25].

- The hyperparameter optimization can be accelerated by the extrapolation of learning curves [9].

- CoDeepNEAT [29] is a evolutionary approach for hyperparameter optimization.

- Gradient-based relaxation of the search, which allows a gradient-based optimization [26], [24], and [40].

Companies invest today massively in hyperparameter optimization for deep learning. A recent result from Google Brain [43] employs reinforcement learning to train a Long Short-Term Memory (LSTM) architecture generator to build a language that describes new deep learning architectures. The LSTM is trained via a policy gradient method to maximize the generation of new architectures. The research explores this method to generate CNNs.

Another recent paper from Google Brain [34] employs an evolutionary algorithm using mutation operators that are inspired by "rules of thumb" coming from various deep learning papers. The evolutionary algorithm uses repeated pairwise competitions of random individuals, and selects from the pair the better performing individual. These Google results show that Google is also looking for unconventional approaches, using their impressive computing capabilities.

## 3 Software for Hyperparameter Optimization

Currently, there are many software packages which include hyperparameter optimization methods. The user can choose between several methods (sometimes called *solvers*), or continue with the default one. Popular packages in this category are [3]:

- LIBSVM [8], BayesianOptimization[2], Spearmint[3], and pyGPGO[4].

- Hyperopt-sklearn provides automatic algorithm configuration of the scikit-learn library. It can be used for both model selection and hyperparameter optimization. Hyperopt-sklearn has the following implemented solvers [6]: RS, SA, and TPE.

- Optunity[5], a Python library containing various optimizers for hyperparameter tuning: GS, RS, PSO, NM, Covariance Matrix Adaptation Evolutionary Strategy, TPE, and Sobol sequences.

Following the idea of Auto-Weka [19], several automated machine learning tools (AutoML) were recently developed. Their ultimate goal is to automatize the end-to-end process of applying machine learning to real-world problems, even for people with no major expertise in this field. Examples of AutoML tools are: MLBox, auto-sklearn, Tree-Based Pipeline Optimization Tool, H2O, AutoKeras, and TransmogrifAI.

Some top companies offer commercial cloud-based AutoML services:

- Google Cloud AutoML[6], enables developers with limited machine learning expertise to train high-quality models specific to their business needs.

- Microsoft Azure ML[7], which can be used to streamline the building, training, and deployment of machine learning models.

- Amazon SageMaker Automatic Model Tuning[8], which can launch multiple training jobs, with different hyperparameter combinations, based on the results of completed training jobs.

In our experiments, we use the Optunity and Hyperopt packages.

---

[2] https://github.com/fmfn/BayesianOptimization
[3] https://github.com/HIPS/Spearmint
[4] https://github.com/hawk31/pyGPGO
[5] http://optunity.readthedocs.io/en/latest/
[6] https://cloud.google.com/automl/
[7] https://docs.microsoft.com/en-us/azure/machine-learning/studio-module-reference/tune-model-hyperparameters
[8] https://aws.amazon.com/sagemaker/

## 4 The WRS Method

This section describes the WRS algorithm and offers some insight into the intuition behind the proposed approach. Full details can be found in [11]. The code is publicly available on GitHub[9].

In the RS approach, hyperparameter optimization translates into the optimization of a objective function $F$ of $d$ variables by generating random values for its parameters and evaluating the function for each of these values [5]. The function computes some quality measure or score of the model (e.g., accuracy), and the variables correspond to the hyperparameters. The assumption is that we want to maximize $F$ by executing a given number $N$ of trials (steps).

RS generates random combinations of hyperparameter values at each iteration $k$, by generating independent random values for each of the $d$ dimensions, each in its own range of values, according to a probability distribution $P_i$ $(i = 1, \ldots, d)$, known beforehand. WRS generates the values for each dimension using a probability of change $p_i$ $(i = 1, \ldots, d)$: the current best value for each dimension $i$ is used with probability $1 - p_i$.

If $(x_1^{k+1}, x_2^{k+1}, \ldots, x_{d-1}^{k+1}, x_d^{k+1})$ is the combination of hyperparameter values evaluated at step $k+1$ in the WRS algorithm, we define random variable $X^{k+1}$ with values $(x_1^{k+1}, x_2^{k+1}, \ldots, x_{d-1}^{k+1}, x_d^{k+1})$ and probabilities $1 = p_1 > p_2, \ldots, > p_{d-1} > p_d$:

$$\begin{cases} (x_1^{k+1}, x_2^{k+1}, \ldots, x_{d-1}^{k+1}, x_d^{k+1}), & \text{with probability } p_d \\ (x_1^{k+1}, x_2^{k+1}, \ldots, x_{d-1}^{k+1}, x_d^k), & \text{with probability } p_{d-1} - p_d \\ \ldots \\ (x_1^{k+1}, x_2^{k+1}, \ldots, x_{d-1}^k, x_d^k), & \text{with probability } p_2 - p_3 \\ (x_1^{k+1}, x_2^k, \ldots, x_{d-1}^k, x_d^k), & \text{with probability } 1 - p_2 \end{cases} \quad (4)$$

A value $x_i^k$ in the definition of $X^{k+1}$ means that $x_i^{k+1} = x_i^k$ and for dimension $i$ we do not generate a new value at step $k + 1$.

The main idea behind the WRS algorithm is that a subset of candidate values that already produced a good result has the potential, in combination with new values for the remaining dimensions, to lead to better values of the objective function. Instead of always generating new values (like in RS), the WRS algorithm uses for a certain number of dimensions the so far best obtained values. The exact number of dimensions that actually change at each iteration is controlled by the probabilities of change assigned to each dimension.

Since we want WRS to have a good coverage of the variation of the objective function, we determine the importance (the weight) of each hyperparameter by computing the variation of the objective function. In order to do this, we first gather some test values, by evaluating function $F$ for a set of randomly chosen inputs. This is basically equivalent to running RS for a predefined number of steps $N_0$, significantly smaller than $N$. We use the obtained data to run an instance of fANOVA [15], which gives us the weight of the $d$ variables of $F$ with respect to the variation of this function. In order to cover as much as possible the variation of the objective function, we assign a greater probability of change to the variables with greater weight. For a parameter that is responsible for only a small part of the variation of $F$, we consider that keeping a temporary optimum, once identified, has the potential to be a more efficient approach. Hence, the probability of change for such a variable will be small.

Algorithm 1 describes one WRS step, corresponding to a trial, applied to the maximization of function $F$. $X^k$ is the best point identified after $k$ iterations, while $F(X^k)$ is the value of $F$ for this input. The probabilities $p_i (i = 1, \ldots, d)$ are the probabilities of change for each of the variables of $F$, and $k_i (i = 1, \ldots, d)$ are the minimum number of values to be generated for each variable. $P_i(x)(i = 1, \ldots, d)$ are the probability distributions used to generate new values for each of the variables. Our assumption in RS and WRS is that the input variables are statistically independent.

Computing the value of $F$ is usually the most time consuming part of the algorithm. In the specific case of hyperparameters optimization, this involves a trial: a train + a test iteration on the model. To avoid recomputing the function's value for the same point, we check if for each iteration at least one variable changed. This explains why at least one of the values for $p_i$ is set to one. If we chose all $p_i$ to be one, we obtain in particular the classical RS implementation. Therefore, WRS is a generalization of RS.

---
[9]https://github.com/acflorea/goptim/tree/keras

---

**Algorithm 1** A WRS Step - Objective Function Maximization [11]

---

**Input:** $F$; $(X^k, F(X^k))$; $p_i, k_i, P_i(x), i = 1, \ldots, d$
**Output:** $(X^{k+1}, F(X^{k+1}))$
1: Randomly generate $p$, uniform in $(0,1)$
2: **for** $i = 1$ to $d$ **do**
3:     **if** $(p_i \geq p$ or $k \leq k_i)$ **then**
4:         // *either the probability condition is met or more samples are needed*
5:         Generate $x_i^{k+1}$ according to $P_i(x)$
6:     **else**
7:         $x_i^{k+1} = x_i^k$
8:     **end if**
9: **end for**
10: // *usually this is the most time consuming step*
11: Compute $F(X^{k+1})$
12: **if** $F(X^{k+1}) \geq F(X^k)$ **then return** $(X^{k+1}, F(X^{k+1}))$
13: **elsereturn** $(X^k, F(X^k))$
14: **end if**

---

**Algorithm 2** WRS - Objective Function Maximization [11]

---

**Input:** $F$; $N$; $P_i(x), i = 1, \ldots, d$
**Output:** $(X^N, F(X^N))$
1: // *Phase 1 - Run RS*
2: **for** $k = 1$ to $N_0 < N$ **do**
3:     Perform RS step, compute $(X^k, F(X^k))$
4: **end for**
5: // *Intermediate phase, determine input for WRS*
6: Determine the probability of change $p_i, i = 1, \ldots, d$
7: Determine the minimum number of required values $k_i, i = 1, \ldots, d$
8: // *Phase 2 - Run WRS*
9: **for** $k = N_0 + 1$ to $N$ **do**
10:     Perform WRS Step described in Algorithm 1, compute $(X^k, F(X^k))$
11: **end forreturn** $(X^N, F(X^N))$

---

Algorithm 2 describes the entire WRS flow. For a well chosen set of $k_i(i = 1, \ldots, d)$ values, WRS has a greater probability than RS to find the global optimum at any step $n$. In [11], we determined how to obtain such a set. On average, within the same computational budget (number of iterations), WRS identifies the global optimum faster than RS. In other words, on average, WRS converges faster than RS. The theoretical convergence of WRS is proved in [11].

## 5 WRS for CNN Hyperparameter Optimization

CNN architecture optimization is currently an unofficial benchmark for testing hyperparameter optimization methods. We will describe in the following the application of the WRS method on a CNN architecture optimization problem. The main goal is to compare its performance against some of the best know hyperparameters optimization algorithms. The choice of a CNN architecture is justified by the popularity of CNN [31] and the ever increasing interest in identifying optimal CNN architectures[12, 20, 38, 42]. Since a complete train-test cycle for a CNN architecture can be extremely time consuming, identifying better algorithms for this class of problems is a hot topic. When comparing two algorithms, "better" means at least one of the following: *a)* using less trials for a similar score; *b)* improving the score within the same computational budget (same number of trials).

The base architecture of the network used in our experiments is shown in Fig. 1. We attempt to optimize the following hyperparameters:

- The number of convolution layers (C) - an integer value in the set $\{3, 4, 5, 6\}$;

- The number of fully-connected layers (F) - an integer value in the set $\{1, 2, 3, 4\}$;

- The number of output filters for each convolution layer (C1, ..., C6) - an integer value in the range $[100, 1024]$;

- The number of neurons for each fully connected layer (F1, ..., F4) - an integer value in the range $[1024, 2048]$.

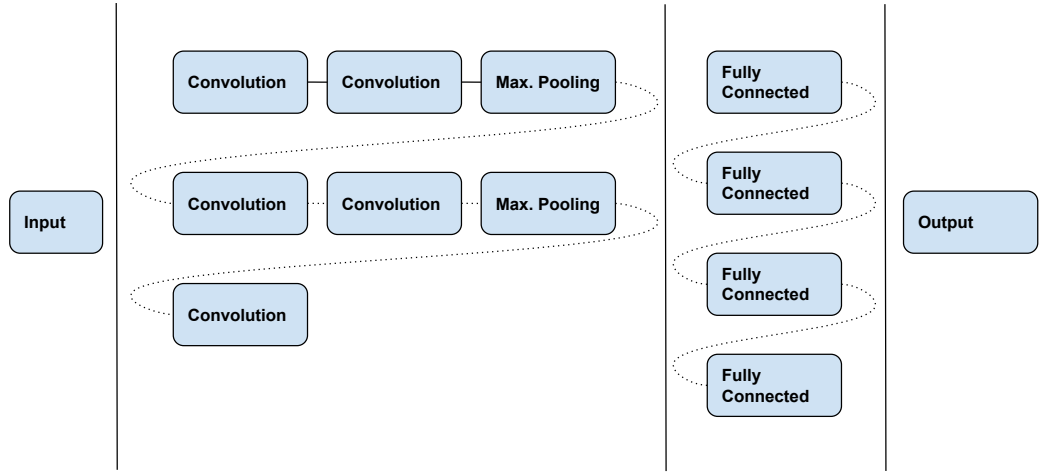

Figure 1: The CNN architecture [11]

On purpose, we use the same problem setup as the one described in [11]. We perform 300 trials for all algorithms, including WRS. We use CIFAR-10 [10] without data augmentation to train the CNN models and compare them in terms of achieved accuracy. The experiments run on an IBM S822LC cluster with IBM POWER8 nodes, NVLink and NVidia Tesla P100 GPUs[11]. On average, the execution time of a trial takes approximately 37 minutes, which leads to a bit less than eight days of running time for each of the 7 tested algorithms.

We compare the obtained results against the following standard techniques:

- Random Search (RS)
- Nelder-Mead (NM)[30]
- Particle Swarm Optimization (PSO)[18]
- Sobol Sequences (SS)[37]
- Bayesian Optimization (BO)[27]
- Tree-structured Parzen Estimator (TPE)[4]

For RS, NM, PSO and SS we use their implementation from the Optunity [44] package, BO comes with it's own implementation, whereas for TPE we use the Hyperopt[7] package.

For WRS we run an initial phase of 110 trials and use the obtained values to rank the hyperparameters and determine the probability of change for each hyperparameter. The obtained values are listed in Table 1.

Table 1: Parameter weights (first line) and probabilities (second line) for CNN[11]

| C | F | C1 | C2 | C3 | C4 | C5 | C6 | F1 | F2 | F3 | F4 |
|---|---|----|----|----|----|----|----|----|----|----|----|
| 7.4 | 11.85 | 0.51 | 0.79 | 1.62 | 0.73 | 2.26 | 1.26 | 26.28 | 0.87 | 3.22 | 1.75 |
| 0.28 | 0.45 | 0.02 | 0.03 | 0.06 | 0.03 | 0.09 | 0.05 | 1.00 | 0.03 | 0.12 | 0.07 |

Fig. 2 shows the least squares five degree polynomial fit on the accuracy results obtained for each of the 300 trials using: WRS, RS, NM, PSO, SS, BO, and TPE. The plot depicts the values obtained at each iteration.

Table 2 lists the best result obtained by each of the tested algorithms, the average accuracy, and the standard deviation across all 300 iterations and (between parenthesis) across the last 100 iterations.

---

[10] http://www.cs.toronto.edu/ kriz/cifar.html
[11] http://www.cwu.edu/faculty/turing-cwu-supercomputer

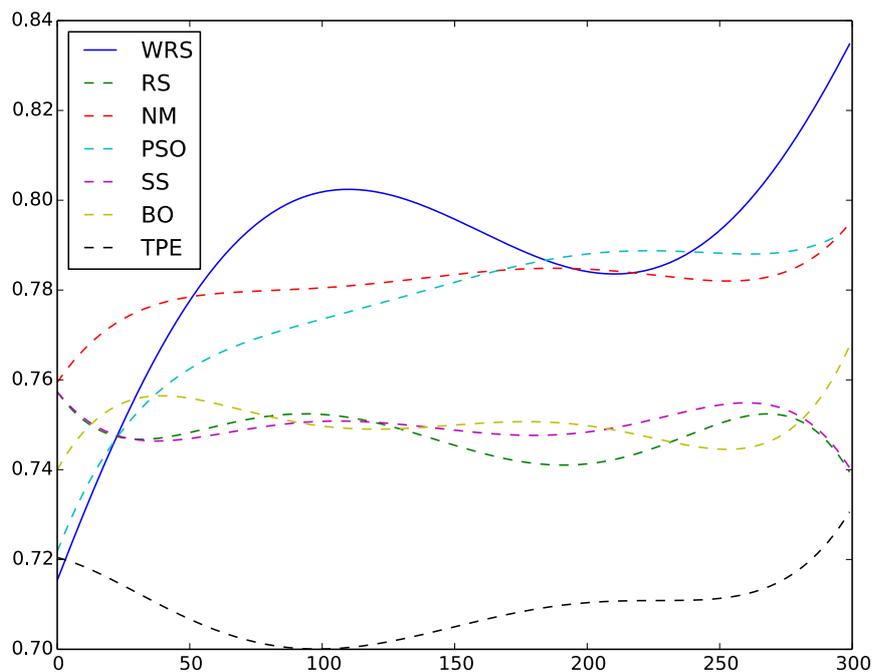

Figure 2: Least squares five degree polynomial fit on RS, NM, PSO, SS, BO, TPE vs. WRS accuracy for CIFAR-10 on 300 trials. The plot shows the values obtained at each iteration

Table 2: Obtained CNN accuracy on CIFAR-10

| Optimizer | Best Result | Average | SD |
|---|---|---|---|
| WRS | **0.85(0.85)** | **0.79(0.79)** | 0.09(0.09) |
| RS | 0.81(0.81) | 0.75(0.8) | 0.04(0.04) |
| NM | 0.81(0.81) | 0.77(0.77) | 0.03(0.03) |
| PSO | 0.83(0.82) | 0.78(0.79) | 0.03(0.03) |
| SS | 0.82(0.82) | 0.75(0.72) | 0.05(0.04) |
| BO | 0.83(0.82) | 0.75(0.75) | 0.04(0.03) |
| TPE | 0.81(0.79) | 0.71(0.71) | 0.04(0.04) |

The CNN architecture that yields the best accuracy, according to WRS, has six convolution layers and one fully-connected layer, with $C1 = 736, C2 = 508, C3 = 664, C4 = 916, C5 = 186, C6 = 352$ and $F1 = 1229$.

According to Table 2, WRS clearly outperforms all other considered algorithms with respect to accuracy, within the same computational budget (number of trials). Among the tested algorithms, WRS yields the highest standard deviation of the results across all 300 iterations, which means that it covers better the variation of the objective function than the other algorithms. This validates our greedy hyperparameter selection procedure used in the WRS algorithm.

From Fig. 2 it can be observed that, in case of WRS, the accuracy has the tendency to improve as the number of iterations increases. This validates the hypothesis that using the best known value for some of the hyperparameters, instead of generating a new one at each iteration, tends to achieve increasingly better results. The same hypothesis is validated by looking at the results for the last 100 iterations in Table 2.

We also notice from Fig. 2 that, whereas for some of the baseline algorithms (e.g., RS, SS, BO) the accuracy seems to decrease or have little variation as the number of iterations increases, the accuracy of the WRS algorithm has the tendency to increase with an increasing number of iterations.

# 6    Conclusions and Open Problems

According our experimental results, the WRS optimization algorithm outperforms several state-of-the art hyperparameter optimization methods. Since we made our code publicly available, we hope that further experiments on other architectures will confirm these results.

However, we are aware that determining the optimal architecture for machine learning models has an additional challenge: the architecture differs for each dataset and therefore requires adjustments for each individual case. Does this mean that in general we cannot optimize the hyperparameters of a given model? For some given problems, recent automatic methods have produced results exceeding those accomplished by human experts [4, 36]. We should be optimistic, but we should be also aware of the scalability limits, since in the most general case, we are dealing with a computationally hard optimization problem.

# References

[1] Albelwi, S.; Mahmood, A. (2016). Analysis of instance selection algorithms on large datasets with deep convolutional neural networks. In *2016 IEEE Long Island Systems, Applications and Technology Conference (LISAT)*, 1–5, 2016.

[2] Albelwi, S.; Mahmood, A. (2017). A framework for designing the architectures of deep convolutional neural networks. *Entropy*, 19(6), 2017.

[3] Andonie, R. (2019). Hyperparameter optimization in learning systems. *Journal of Membrane Computing*, 2019.

[4] Bergstra, J.; Bardenet, R.; Bengio, Y.; Kégl, B. (2011a). Algorithms for Hyper-Parameter Optimization. In *25th Annual Conference on Neural Information Processing Systems (NIPS 2011)*, volume 24 of *Advances in Neural Information Processing Systems*, Granada, Spain. Neural Information Processing Systems Foundation, 2011.

[5] Bergstra, J.; Bardenet, R.; Bengio, Y.; Kégl, B. (2011b). Algorithms for hyper-parameter optimization. In Shawe-Taylor, J., Zemel, R. S., Bartlett, P. L., Pereira, F. C. N., and Weinberger, K. Q., editors, *NIPS*, 2546–2554, 2011.

[6] Bergstra, J.; Komer, B.; Eliasmith, C.; Yamins, D.; Cox, D. D. (2015). Hyperopt: a Python library for model selection and hyperparameter optimization. *Computational Science and Discovery*, 8(1), 014008, 2015.

[7] Bergstra, J.; Yamins, D.; Cox, D. D. (2013). Making a science of model search: Hyperparameter optimization in hundreds of dimensions for vision architectures. In *Proceedings of the 30th International Conference on International Conference on Machine Learning - Volume 28*, ICML'13, JMLR.org, I–115–I–123, 2013.

[8] Chang, C.-C.; Lin, C.-J. (2011). LIBSVM: A library for support vector machines. *ACM Transactions on Intelligent Systems and Technology*, 2:27:1–27:27, 2011.

[9] Domhan, T.; Springenberg, J. T.; and Hutter, F. (2015). Speeding up automatic hyperparameter optimization of deep neural networks by extrapolation of learning curves. In *Proceedings of the 24th International Conference on Artificial Intelligence*, IJCAI'15, AAAI Press, 3460–3468, 2015.

[10] Florea, A. C.; Andonie, R. (2018). A dynamic early stopping criterion for random search in svm hyperparameter optimization. In Iliadis, L., Maglogiannis, I., and Plagianakos, V., editors, *Artificial Intelligence Applications and Innovations*, Cham. Springer International Publishing, 168–180, 2018.

[11] Florea, A.-C.; Andonie, R. (2019). Weighted random search for hyperparameter optimization. *International Journal of Computers Communications & Control*, 14(2), 154–169, 2019.

[12] He, K.; Zhang, X.; Ren, S.; Sun, J. (2016). Deep residual learning for image recognition. *2016 IEEE Conference on Computer Vision and Pattern Recognition (CVPR)*, 770–778, 2016.

[13] Hinz, T.; Navarro-Guerrero, N.; Magg, S.; Wermter, S. (2018). Speeding up the hyperparameter optimization of deep convolutional neural networks. *International Journal of Computational Intelligence and Applications*, 17(02), 1850008, 2018.

[14] Hutter, F.; Hoos, H.; Leyton-Brown, K. (2014a). An efficient approach for assessing hyperparameter importance. In *Proceedings of the 31th International Conference on Machine Learning, ICML 2014, Beijing, China, 21-26 June 2014*, 754–762, 2014.

[15] Hutter, F.; Hoos, H.; Leyton-Brown, K. (2014b). An efficient approach for assessing hyperparameter importance. In *Proceedings of International Conference on Machine Learning 2014 (ICML 2014)*, 754–762, 2014.

[16] Ilievski, I.; Akhtar, T.; Feng, J.; Shoemaker, C. A. (2016). Hyperparameter optimization of deep neural networks using non-probabilistic RBF surrogate model. *CoRR*, abs/1607.08316, 2017.

[17] Ilievski, I.; Akhtar, T.; Feng, J.; Shoemaker, C. A. (2017). Efficient hyperparameter optimization for deep learning algorithms using deterministic RBF surrogates. In *Proceedings of the Thirty-First AAAI Conference on Artificial Intelligence, February 4-9, 2017, San Francisco, California, USA.*, 822–829, 2017.

[18] Kennedy, J.; Eberhart, R. C. (1995). Particle swarm optimization. In *Proceedings of the IEEE International Conference on Neural Networks*, pages 1942–1948, 1995.

[19] Kotthoff, L.; Thornton, C.; Hoos, H. H.; Hutter, F.; Leyton-Brown, K. (2017). Auto-WEKA 2.0: Automatic model selection and hyperparameter optimization in WEKA. *Journal of Machine Learning Research*, 18(25), 1–5, 2017.

[20] Krizhevsky, A.; Sutskever, I.; Hinton, G. E. (2012). Imagenet classification with deep convolutional neural networks. In *Proceedings of the 25th International Conference on Neural Information Processing Systems - Volume 1*, NIPS'12, pages 1097–1105, USA. Curran Associates Inc., 2012.

[21] Larson, J.; Menickelly, M.; Wild, S. M. (2019). Derivative-free optimization methods. *Acta Numerica*, 28, 287–404, 2019.

[22] Lemley, J.; Jagodzinski, F.; Andonie, R. (2016). Big holes in big data: A monte carlo algorithm for detecting large hyper-rectangles in high dimensional data. In *2016 IEEE 40th Annual Computer Software and Applications Conference (COMPSAC)*, 1, 563–571, 2016.

[23] Li, L.; Jamieson, K. G.; DeSalvo, G.; Rostamizadeh, A.; Talwalkar, A. (2016). Efficient hyperparameter optimization and infinitely many armed bandits. *CoRR*, abs/1603.06560, 2016.

[24] Liu, H.; Simonyan, K.; Yang, Y. (2018). DARTS: differentiable architecture search. *CoRR*, abs/1806.09055. 2018.

[25] Loshchilov, I.; Hutter, F. (2016). CMA-ES for hyperparameter optimization of deep neural networks. *CoRR*, abs/1604.07269, 2016.

[26] Luo, R.; Tian, F.; Qin, T.; Chen, E.; Liu, T.-Y. (2018). Neural architecture optimization. In Bengio, S., Wallach, H., Larochelle, H., Grauman, K., Cesa-Bianchi, N., and Garnett, R., editors, *Advances in Neural Information Processing Systems 31*, Curran Associates, Inc, 7816–7827, 2018.

[27] Martinez-Cantin, R. (2014). Bayesopt: A bayesian optimization library for nonlinear optimization, experimental design and bandits. *Journal of Machine Learning Research*, 15, 3915–3919, 2014.

[28] Miikkulainen, R.; Liang, J.; Meyerson, E.; Rawal, A.; Fink, D.; Francon, O.; Raju, B.; Shahrzad, H.; Navruzyan, A.; Duffy, N.; Hodjat, B. (2019). Chapter 15 - evolving deep neural networks. In Kozma, R., Alippi, C., Choe, Y., and Morabito, F. C., editors, *Artificial Intelligence in the Age of Neural Networks and Brain Computing*, Academic Press, 293 – 312, 2019.

[29] Miikkulainen, R.; Liang, J. Z.; Meyerson, E.; Rawal, A.; Fink, D.; Francon, O.; Raju, B.; Shahrzad, H.; Navruzyan, A.; Duffy, N.; Hodjat, B. (2017). Evolving deep neural networks. *CoRR*, abs/1703.00548, 2017.

[30] Nelder, J. A.; Mead, R. (1965). A Simplex Method for Function Minimization. *Computer Journal*, 7, 308–313, 1965.

[31] Patterson, J.; Gibson, A. (2017). *Deep Learning: A Practitioner's Approach*. O'Reilly Media, Inc., 1st edition, 2017.

[32] Raschka, S. (2018). Model evaluation, model selection, and algorithm selection in machine learning. *CoRR*, abs/1811.12808, 2018.

[33] Real, E.; Aggarwal, A.; Huang, Y.; Le, Q. V. (2018). Regularized evolution for image classifier architecture search. *CoRR*, abs/1802.01548, 2018.

[34] Real, E.; Moore, S.; Selle, A.; Saxena, S.; Suematsu, Y. L.; Le, Q. V.; Kurakin, A. (2017). Large-scale evolution of image classifiers. *CoRR*, abs/1703.01041, 2017.

[35] Shahriari, B.; Swersky, K.; Wang, Z.; Adams, R. P.; de Freitas, N. (2016). Taking the human out of the loop: A review of Bayesian optimization. *Proceedings of the IEEE*, 104(1), 148–175, 2016. ,

[36] Snoek, J.; Larochelle, H.; Adams, R. P. (2012). Practical Bayesian optimization of machine learning algorithms. In *Advances in Neural Information Processing Systems 25: 26th Annual Conference on Neural Information Processing Systems 2012. Proceedings of a meeting held December 3-6, 2012, Lake Tahoe, Nevada, United States.*, 2960–2968, 2012.

[37] Sobol, I. (1976). Uniformly distributed sequences with an additional uniform property. *USSR Computational Mathematics and Mathematical Physics*, 16(5), 236 – 242, 1976.

[38] Szegedy, C.; Liu, W.; Jia, Y.; Sermanet, P.; Reed, S.; Anguelov, D.; Erhan, D.; Vanhoucke, V.; Rabinovich, A. (2015). Going deeper with convolutions. In *Computer Vision and Pattern Recognition (CVPR)*, 2015.

[39] Talathi, S. S. (2015). Hyper-parameter optimization of deep convolutional networks for object recognition. In *2015 IEEE International Conference on Image Processing (ICIP)*, 3982–3986, 2015.

[40] Wu, B.; Dai, X.; Zhang, P.; Wang, Y.; Sun, F.; Wu, Y.; Tian, Y.; Vajda, P.; Jia, Y.; Keutzer, K. (2018). Fbnet: Hardware-aware efficient convnet design via differentiable neural architecture search. *CoRR*, abs/1812.03443, 2018.

[41] Young, S. R.; Rose, D. C.; Karnowski, T. P.; Lim, S.-H.; Patton, R. M. (2015). Optimizing deep learning hyper-parameters through an evolutionary algorithm. In *Proceedings of the Workshop on Machine Learning in High-Performance Computing Environments*, MLHPC '15, New York, NY, USA. ACM, 4, 1–5, 2015.

[42] Zeiler, M. D.; Fergus, R. (2014). Visualizing and understanding convolutional networks. In Fleet, D., Pajdla, T., Schiele, B., and Tuytelaars, T., editors, *Computer Vision – ECCV 2014*, Cham. Springer International Publishing, 818–833, 2014.

[43] Zoph, B.; Le, Q. V. (2016). Neural architecture search with reinforcement learning. *CoRR*, abs/1611.01578, 2016.

[44] [online] Optunity. Available: http://optunity.readthedocs.io/en/latest/. Accessed: 2017-09-01.

*Cite this paper as:*

Andonie, R.; Florea, A.-C.(2020). Weighted Random Search for CNN Hyperparameter Optimization, *International Journal of Computers Communications & Control*, 15(2), 3868, 2020.

https://doi.org/10.15837/ijccc.2020.2.3868